\documentclass[fleqn,10pt]{wlscirep}
\usepackage[utf8]{inputenc}
\usepackage[T1]{fontenc}
\usepackage{graphicx}
\usepackage{multirow}

\title{Analysing Wideband Absorbance Immittance in Normal and Ears with Otitis Media with Effusion Using Machine Learning}

\author[1,+]{Emad M. Grais}
\author[2,+]{Xiaoya Wang}
\author[3,4,+]{Jie Wang}
\author[1,*]{Fei Zhao}
\author[5]{Wen Jiang}
\author[6,7]{Yuexin Cai}
\author[3,4]{Lifang Zhang}
\author[2]{Qingwen Lin}
\author[6,7,*]{Haidi Yang}
\affil[1]{Cardiff Metropolitan University, Centre for Speech and Language Therapy and Hearing Science, School of Sport and Health Sciences,  Cardiff, CF5 2YB, UK}
\affil[2]{Guangzhou Women and Children Medical Centre, Department of Otolaryngology, Guangzhou City, Guangdong Province, China}
\affil[3]{Department of Otolaryngology Head and Neck Surgery, Beijing Tongren Hospital, Capital Medical University, Key Laboratory of Otolaryngology Head and Neck Surgery, Ministry of Education, Beijing Engineering Research Center of Hearing Technology, Beijing, 100730, China}
\affil[4]{Key Laboratory of Otolaryngology Head and Neck Surgery, Ministry of Education, Beijing Engineering Research Centre of Hearing Technology, Beijing, 100730, China}
\affil[5]{Xuzhou Medical University, Department of Hearing and Speech Sciences, Xuzhou City, Jiangsu Province, China}
\affil[6]{Sun Yat-sen University, Sun Yat-sen Memorial Hospital, Department of Otolaryngology, Guangzhou City, Guangdong Province, China}
\affil[7]{Sun Yat-sen University, Institute of Hearing and Speech-Language Science, Guangzhou City, Guangdong Province, China}
\affil[*]{Corresponding authors Fei Zhao: fzhao@cardiffmet.ac.uk and Haidi Yang: yanghd@mail.sysu.edu.cn}
\affil[+]{These authors contribute equally: Xiaoya Wang and Jie Wang in data collection, and Emad Grais in data analysis}


\begin{abstract}
Wideband Absorbance Immittance (WAI) has been available for more than a decade, however its clinical use still faces the challenges of limited understanding and poor interpretation of WAI results. This study aimed to develop Machine Learning (ML) tools to identify the WAI absorbance characteristics across different frequency-pressure regions in the normal middle ear and ears with otitis media with effusion (OME) to enable diagnosis of middle ear conditions automatically. Data analysis including pre-processing of the WAI data, statistical analysis and classification model development, together with key regions extraction from the 2D frequency-pressure WAI images are conducted in this study. Our experimental results show that ML tools appear to hold great potential for the automated diagnosis of middle ear diseases from WAI data. The identified key regions in the WAI provide guidance to practitioners to better understand and interpret WAI data and offer the prospect of quick and accurate diagnostic decisions. 
\end{abstract}
\begin{document}

\flushbottom
\maketitle
%
%
\thispagestyle{empty}


\section{Introduction}
\label{Sec:intro}
The human middle ear functions importantly for effective sound transmission by playing an impedance matching device between the low impedance of air and high impedance of cochlear fluids \cite{Zhao:09:feametfrp}. Tympanometry is a useful tool to measure acoustic admittance changes of the middle ear system as air pressure is varied in the external ear canal \cite{Margolis:02:tbpca}. Conventional tympanometry with a single low-frequency (usually 220 or 226 Hz) probe tone is used routinely in audiological and otological assessment. Over the past five decades, a large body of research evidence has shown that tympanometry is an essential tool to detect certain types of middle ear pathologies in ENT/Audiology clinics \cite{Margolis:02:tbpca}. Technological advances in the field of assessments of middle ear function have expanded the frequency range from single probe tones (226/1000 Hz) to multiple frequency measurements delivered as a sweep through a series of frequencies \cite{Shahnaz:97:smtnoe}. This multiple frequency tympanometry (MFT) has been shown to provide better sensitivity and specificity in the detection of some middle ear pathologies, such as otosclerosis and ossicular discontinuity \cite{Zhao:02:medcpo}.
In recent years a commercialised MFT device that uses Wideband Absorbance Immittance (WAI), also known as Wideband Energy Reflectance (WBER) or Otoreflectance has been developed \cite{Margolis:99:wrtna}. It is a system designed to assess wideband acoustic transfer functions of the middle ear over a wide frequency range from 0.25 to 8.0 kHz \cite{Liu:08:watpssdranh}. The acoustic absorbance characterises the ratio of the absorbed sound energy to the incident sound energy.  A number of studies have shown that measurement of absorbance has several advantages over traditional tympanometry \cite{Margolis:99:wrtna,Zhao:07:tumemocer}. Measurement of WAI is simple, fast, objective, reproducible and non-invasive, and because some changes of energy absorbance are associated with certain types of middle ear pathologies, the WAI has unique features that provide important diagnostic information of patients with middle ear disorders. Keefe et al. \cite{Keefe:12:waaapchlc} found that the likelihood-ratio predictors for wideband absorbance at ambient and tympanometric pressure was higher (0.97 to 0.93) than the predictors for conventional 226Hz tympanometry (0.80 to 0.93) in the detection of conductive hearing loss. In addition  Keefe and Simon \cite{Keefe:03:etpchloca} showed that the sensitivity for diagnosing childhood otitis media with effusion (OME) increased from 27\% to 78\% when adding  WAI measurement to the test battery thereby reducing inappropriate diagnosis and costs associated with the condition.

Recent studies have also proven the significant advantages of WAI in providing additional information on middle ear function using a wider frequency range as a function of pressure, e.g., at ambient pressure and peak pressure,  plotted in  two- and three-dimensional graphs \cite{Liu:08:watpssdranh, Wang:18:3diawtneome, Hougaard:20:swtadcnha}. The multidimensional graphs obtained from WAI enable the clinician to better understand the dynamic characteristics of the middle ear by recognizing specific tympanometric patterns associated with middle ear pathologic change. Niemczyk et al. \cite{Niemczyk:19:wtamoe} investigated WAI patterns in ears with intraoperatively confirmed otosclerosis by analysing resonance frequency, and number of peaks with detailed descriptions in terms of height and width. Although the patterns were statistically significantly different and provided important diagnostic information in terms of otosclerotic status, this approach appears less helpful clinically for the purpose of differential diagnosis as the absorbance patterns overlap with absorbance graphs found in the normal ear condition and other middle ear disorders, for example there is a similar characteristic of significantly reduced absorbance in frequencies below 2000 Hz in cases of OME \cite{Wang:18:3diawtneome}. Therefore the main challenges clinicians still face are to understand, interpret and use WAI data as an effective and accurate diagnostic tool in ENT and Audiology clinics. 

In addition, there is little research being undertaken to investigate the use of whole WAI data on energy absorbance, within which is embedded substantial information associated with energy transfer function of the middle ear, particularly in the high frequency region under various middle ear pressures \cite{Liu:08:watpssdranh,Hougaard:20:swtadcnha}. A recent study by Hougaard et al. \cite{Hougaard:20:swtadcnha} analysed the wideband energy absorbance (EA) tympanogram from 99 ears in normal middle ear conditions. The results revealed a trend of increasing EA in the lower frequencies as a function of frequency regardless of ear pressure. EA peaked at around 4.0–5.0 kHz under positive pressures between +50 and +150 daPa, followed by a sharp decrease at higher frequencies. Although this study provided important 3D absorbance information in adults with normal hearing and middle ear function, future studies are necessary to further understand the potential for WAI in clinical applications.

Machine learning (ML) tools have been used to explore and process different data to extract useful information, make predictions and inform decision making \cite{Han:12:dmct, Goodfellow-et-al-2016}. The initial motivation underlying this study is to address questions that have arisen from clinical challenges and fill in gaps in the literature, particularly in relation to the limited understanding and poor interpretation of WAI results across different pressures in various frequency regions. Therefore, we aimed to identify the characteristics of WAI absorbance across different frequency-pressure regions in normal middle ear conditions and ears with OME and to develop ML tools to automatically diagnose ears as normal or with OME. Initial statistical analysis compared absorbance values at different frequency-pressure regions in normal middle ears and middle ears with OME. This was followed by an evaluation of the performance of five ML models in classifying the WAI data as being from a normal middle ear conditions or ear with OME. We also aimed to extract the key regions in the WAI data that could support the clinicians in deciding whether the WAI data was from a normal ear or ear with OME. To the best of our knowledge, the present study is the first to use ML tools to better understand and interpret WAI results, providing automated diagnosis of ears with OME and further facilitate its clinical application.  

\section{Materials and Methods}
\label{Sec:MM}
\subsection{Materials}
\label{Sec:Materials}
\subsubsection*{Wideband Absorbance Immittance (WAI) data acquisition}
\label{Sec:WAI_acq}
A total of 672 sets of WAI data were collected from patients and volunteers in five hospitals in Beijing, Guangzhou and Xuzhou, China. 423 had normal middle ear function and 249 had OME. The definition and inclusion criteria for normal middle ear and ears with otitis media with effusion were:- 
\begin{itemize}
\item \textbf{\textit{Normal middle ear function}}:
1) No history of inflammation or disease that has impacted the middle ear, nor any recent hearing disability and aural symptoms; 2) Otoscopy: normal tympanic membrane lustre  normal, no atrophy, scar, retraction or perforation; 3) Tympanometry:  normal range of middle ear pressures within -50 to +50 daPa (adults) or -100 to +50 daPa (children). Peak compliance range from 0.3 to 1.4 ml (adults) or 0.3 to 0.9 ml (children) (Type A tympanogram) 
%
%
\item \textbf{\textit{Otitis media with effusion (OME)}}:
1) OME is defined as fluid of varying amount and viscosity accumulated in the middle ear as a result of Eustachian tube dysfunction; 2) Otoscopy: tympanic membrane lustre  is dull, either with air bubbles or a fluid line or  retraction; 3) Tympanometry: abnormality of middle ear function measured and diagnosed by conventional tympanometry, showing a) significant negative middle ear pressure in the presence of normal static compliance (Type C tympanogram), i.e., range of middle ear pressure is less than -50 daPa (adults) or -100 daPa (children), and range of peak compliance from 0.3 to 1.4 ml (adults) or 0.3 to 0.9 ml (children), or b) no measureable middle ear pressure or static compliance (Type B tympanogram), i.e., a flat trace. 
\end{itemize}
\subsection{Methods}
\label{Sec:methods}
The work included pre-processing of the WAI data, statistical analysis and ML model development, together with further significant region extraction from the 2D frequency-pressure WAI images.
\subsubsection{WAI measuring system and data pre-processing}
\label{Sec:wai_system}
\begin{figure}[ht]
\centering
\includegraphics[width=\linewidth]{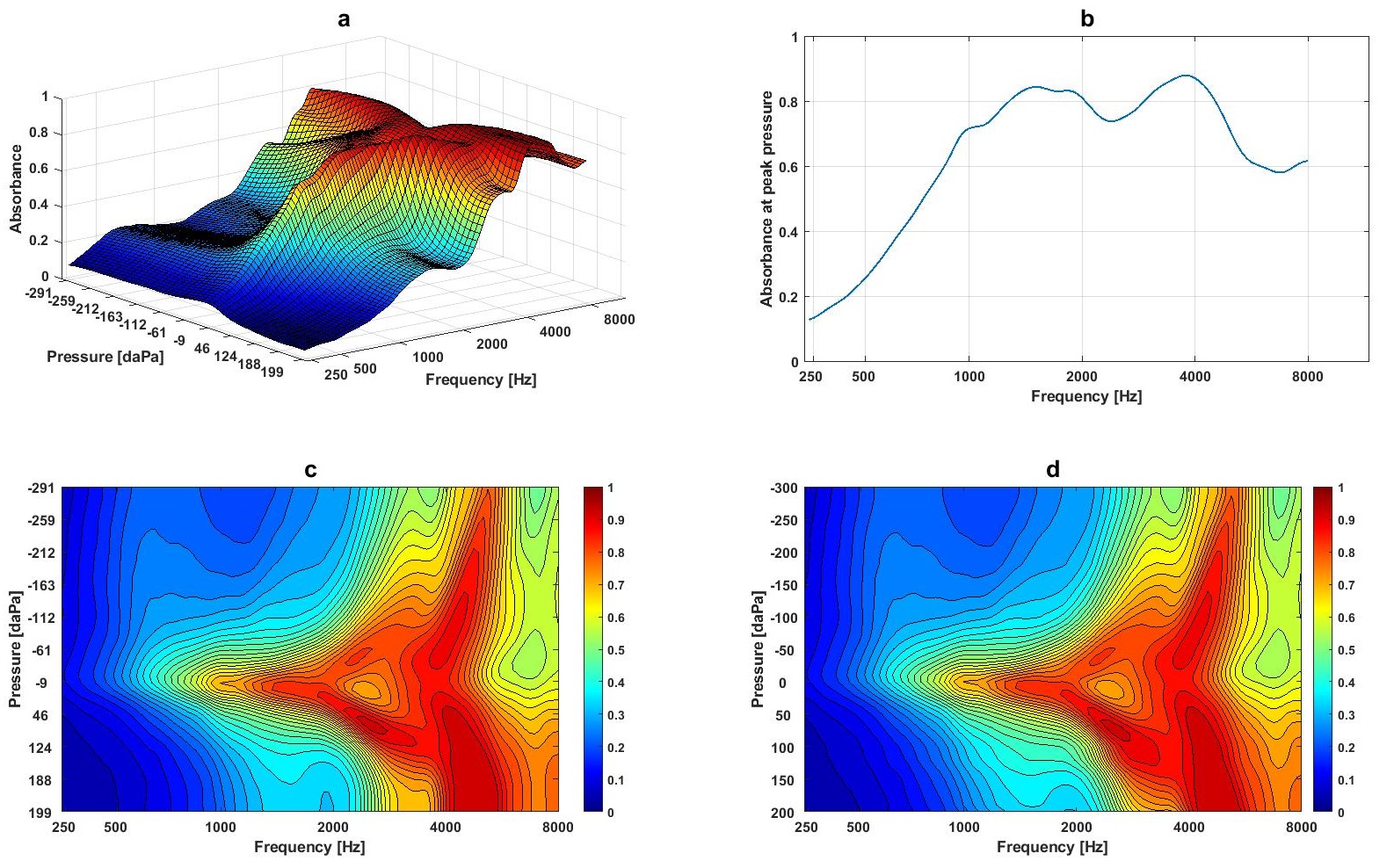}
\caption{\small{An example of the 3D WAI image and data pre-processing:  
a) An example of 3D WAI data obtained from a participant with normal middle ear function aged 22;  
b) the 2D frequency-absorbance plot at the peak pressure obtained from same participant;  
c) the 2D frequency-pressure image converted from the 3D WAI in Figure \ref{fig1}a;  
d) the 2D frequency-pressure image converted from the 3D WAI in Figure \ref{fig1}a after interpolating the pressure values on the Y-axis.
}}
\label{fig1}
\end{figure}

A Titan IMP440 (Interacoustics, Denmark) was used to measure the 3D wideband absorbance. Figure \ref{fig1}a shows an example of a 3D image of Wideband Absorbance Immittance (WAI) obtained from a participant with normal middle ear function aged 22 years. The figure shows one dimension of the WAI to be frequency, the second pressure, and the third absorbance value \cite{Liu:08:watpssdranh}. Frequencies varied from 226 Hz to 8000 Hz with 1/24-octave frequency-intervals. Pressures varied from -300 to +200 daPa, and absorbance was between zero and one. Theoretically, higher absorbance indicates a better transfer function of the middle ear, whereas lower absorbance means less energy being passed through the middle ear, indicating pathological change in the middle ear. Figure \ref{fig1}b shows the absorbance curve at peak pressure across a wide frequency range currently used to evaluation the middle ear function. Figure \ref{fig1}c shows a 2D image using the domains of frequency and pressure corresponding to the WAI data in Figure \ref{fig1}a. Values of absorbance show as values of pixels in the image. There are 107 bins across the frequency axis (X-axis) starting at 226 HZ and rising to 8000 Hz. Because the pressure axis was unevenly sampled due to artefact rejection of noisy samples, ear canal volume, and probe fit \cite{Hougaard:20:swtadcnha}, we resampled the pressure axis between -300 to +200 daPa in 10-daPa steps using the Piecewise Cubic Hermite Interpolating Polynomial \cite{Fritsch:80:mpci}. As a result, there are 51 pressure values on the Y-axis as shown in Figure \ref{fig1}d. There were a total of 5457 data points, i.e., 107-frequency bins $\times$ 51-pressure values in the 2D frequency-pressure WAI images.
\subsubsection{Statistical analysis}
Mean and variance of absorbance at each frequency-pressure region was calculated for participants with normal middle ear function and those with OME. A Wilcoxon test rank sum test was used to determine any significant differences between the normal and OME ears. The level of significance was set at the conventional 5\% level.  
\subsubsection{Development of machine learning (ML) models for WAI classification }
\label{dmlmwaic}
Basic ML classifiers were used to process the 2D WAI images after interpolating the pressure axis; K-nearest neighbours classifier, support vector machines (SVM), and random forest (RF). Deep learning (DL) based classifiers, such as feedforward neural networks (FNN) and convolutional neural networks (CNN) were also examined. The ultimate goal of developing ML classifiers was to enable the automatic classification of WAI data as being from normal or OME ears.
\subsubsection{Extracting the important regions from 2D WAI images}
In clinical practice, as a part of clinical assessment, medical images (e.g., X-ray, CT scan and MRI) play an important role to detect and identify the pathological changes, and thus facilitate clinical diagnosis. Experienced clinicians are usually aware of the key regions to be examined, and consequently achieve a quick and accurate judgement that leads to an appropriate diagnostic decisions. As same as the scan images measured in the healthcare, 2D frequency-pressure WAI images could also play an important role in identifying the specific pathological changes associated with middle ear disorders. However, due to lack of understanding and appropriate interpretations of the WAI measurements, as the first of its kind in using ML to analyse the WAI data, the feature selection tools were used to extract the important regions in the 2D frequency-pressure WAI that provide valuable guidance to audiologists and ENT physicians in their diagnostic decisions. In this study, two feature extraction techniques were conducted, i.e., 1) random forest classifiers as a feature selection tool to extract the important features of the 2D images, and 2) significance tests over the 2D frequency-pressure WAI images to extract regions showing the most significant differences between normal ears and OME ears. 
\section*{Ethics} All methods used in this study were approved by Cardiff School of Sport and Health Sciences Ethical Committee under the Cardiff Metropolitan University ethical guidelines and regulations (Ethical reference number: Sta-3013). Informed consent was obtained from all subjects, and if subjects are under 18, from a parent and/or legal guardian. The anonymous WAI data were analysed when machine learning tools were used. 
\section{Results}
\label{sec:results}
\subsection{WAI characteristics and statistical analysis of the normal middle ear condition compared to ears with OME}
\label{sec:waicsanmecceo}
\begin{figure}[ht]
\centering
\includegraphics[width=\linewidth]{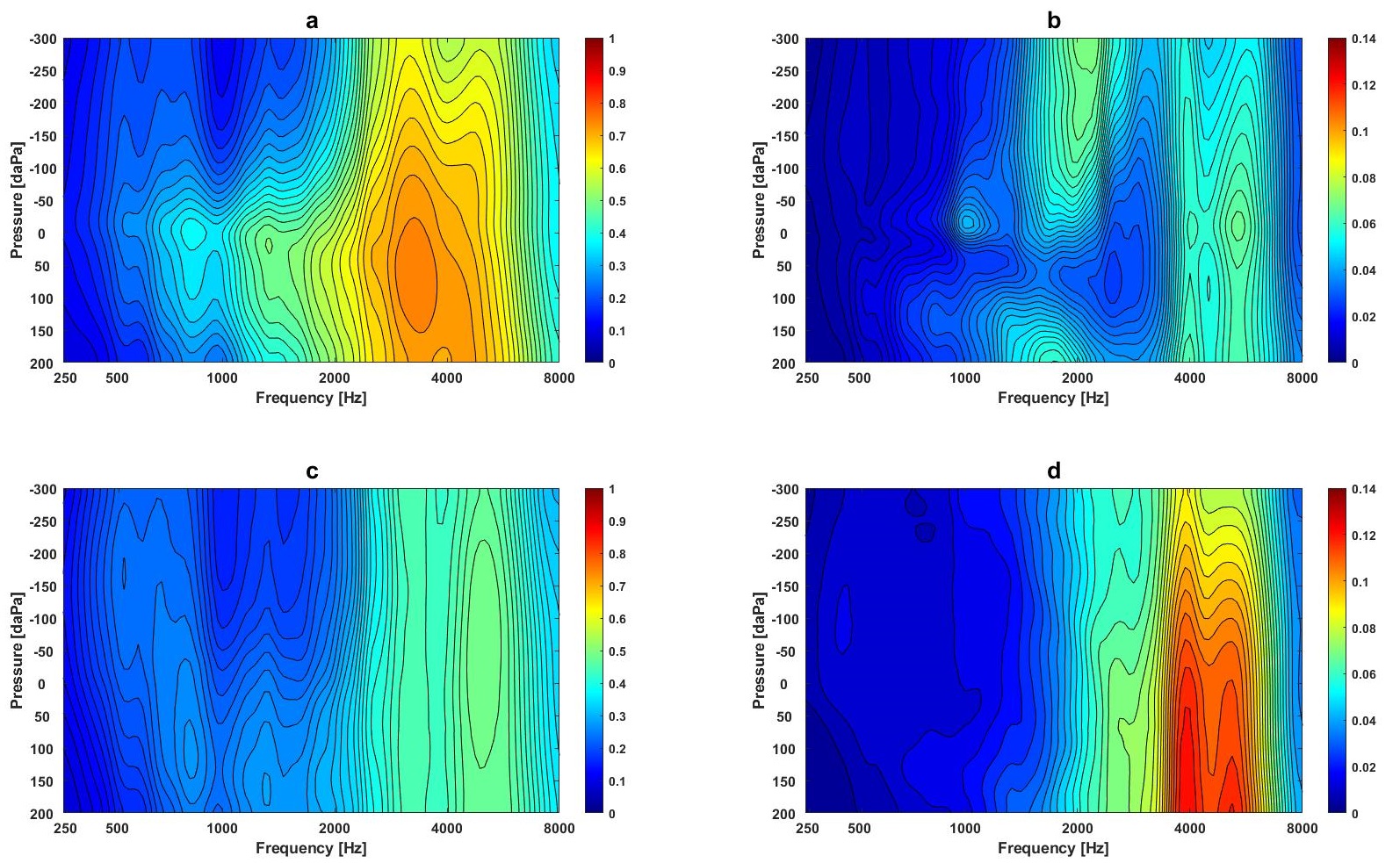}
\caption{\small{The mean and variance in absorbance at different frequency-pressure regions in the normal middle ear condition and ears with OME:
a) Mean absorbance contour plot at different  frequencies and  pressures in the normal middle ear condition;
b) Variance of absorbance  contour plot at different frequencies and pressures in the normal middle ear condition;  
c) Mean of absorbance contour plot at different frequencies and  pressures in ears with OME;
d) Variance of absorbance contour plot at different frequencies and  pressures in ears with OME. 
}}
\label{fig2}
\end{figure}

Figures \ref{fig2}a-d show the mean and variance of absorbance at different frequencies and pressures in the normal middle ear and ears with OME. The averaged absorbance contour for the normal middle ear condition showed a peak area at the centre frequency of 820 Hz at 0 daPa with an absorbance value of 0.39 (frequency range: 771-917 Hz with pressures between -30 and +30 daPa and absorbance value 0.4), and the second peak at the centre frequency of 1335 Hz at +20 daPa with absorbance value of 0.50 (frequency range: 1300-1370 Hz with pressures between 0 and +40 daPa and absorbance value 0.5). The largest peak occurred at the centre frequency 3270 Hz at +65 daPa, with absorbance at 0.76) (frequencies range: 2900-3700 Hz with  pressures between -30 and +160 daPa and  absorbance values between 0.75 and 0.76 (Figure \ref{fig2}a). 
Figure \ref{fig2}b shows the contour of variance for the normal middle ear condition. There were a couple of areas showing larger variances in absorbance obtained from normal middle ears, i.e., variance value of 0.07 at frequencies between1834 and 2370Hz at pressures between -300 and -110daPa, and the variance value of 0.07 at  frequencies between 5180 and 5500 Hz at  pressures between -30 and +10daPa.
In comparison, ears with OME showed the largest peak centred at 5000 Hz with pressure at -30 daPa with an absorbance value of 0.5 (frequency range: 4500 to 5500 Hz,  pressures from -220 to +130 daPa with an absorbance value of 0.5) (Figure \ref{fig2}c). The highest value in the averaged absorbance was significantly lower in the OME ears than the normal ears (0.49 vs. 0.76, $p<0.0005$). Significantly higher variances in  absorbance were found in OME ears, with the largest averaged variance occurring between 3700 Hz and 5500 Hz and at pressures  between +40 and +200 daPa) (frequency range from 3700Hz to 5500Hz, pressures from +40 to +200 daPa and variance values from 0.11 to 0.12 (Figure \ref{fig2}d). Further statistical analysis at each frequency-pressure point showed  88\% of  data points (4782 out of 5457) to have significant differences in absorbance values between normal ears and OME ears (Wilcoxon rank sum test, $Z=3.04$, $p<0.0005$).
\subsection{The outputs derived from various classification models: accuracy, the area under the ROC curve, F1 score, precision and recall for categorising middle ear function as normal or OME}
As indicated in the methods Section \ref{dmlmwaic}, different ML classifiers were developed and examined to determine their ability to categorise middle ear function as normal or OME using the 2D frequency-pressure WAI images. For the KNN, SVM, RF, and FNN classifiers, the 2D images were converted into vectors with the dimension 5457, i.e., 107-frequency bins $\times$ 51-pressure values. The 2D images were used directly for the CNN. Data were normalized to have a zero mean and unit variance for all classifiers. We used the 10-fold cross validation to test each model. There is some randomization in the process of RF and in the initialization of FNN and CNN, and since the performance of these classifiers depends strongly on this initialization, especially with small datasets, we ran the experiments three times with different random initializations for each model run. To implement the neural networks, Keras with TensorFlow backend was used \cite{chollet2015keras}, and for the remaining classifiers, we used Scikit-learn Python based library \cite{scikit-learn}. 
Table \ref{table:classification} summarises the results derived from each ML model in terms of; accuracy, area under the ROC curve, precision, recall/sensitivity and F1-score for predicting normal from OME \cite{Han:12:dmct}. Overall the different ML models produced different outputs. The area under the ROC curve ranged from 0.74 to 0.79, while the accuracy for determining the middle ear function ranged from 0.74 to 0.82. Of these, the CNNs had the highest accuracy as the most promising models.
We tested different KNN classifiers with different number of neighbours in queries, i.e., $K=1, 3, 15$. All points in each neighbourhood were weighted equally and Euclidean distance was used to measure the distance between data samples. The results of KNN shown in Table \ref{table:classification} indicate that increasing $K$ improves the accuracy and precision for OME, the recall for the normal and F1-scores for both classes, but decreases the recall of the OME and slightly decreases the precision for the normal cases. This indicates that with increasing $K$ the model is biased toward classifying the data into normal and this perhaps because the number of normal samples is higher than the number of OME samples in our dataset. For SVM, different kernels were examined, such as, linear, polynomial (Poly), radial basis function (RBF), and sigmoid \cite{Han:12:dmct}. The degree for the polynomial kernel is three. As shown in Table \ref{table:classification}, the accuracy and F1-scores, for the polynomial and RBF kernels were better than the other kernels. 
For the RF, different numbers of decision trees were tested in the forest combination, e.g., 10, 100, 500. The RF was run three times with different randomization for each run. As shown in Table \ref{table:classification}, the best results (accuracy, F1-scores, precision, and recalls for both classes) were obtained using 100 and 500 decision trees in the RF. 
With the experiments using Feedforward neural networks (FNN), different number of layers and different number of nodes in each layer were developed and examined. The detailed structure of each FNN is described in Table \ref{table:models}. The rectified linear unit (ReLU) was used as an activation function in the hidden layers, and the sigmoid activation function was used in the output layer. Dropout value 20\% was used after each hidden layer. The tested FNNs gave almost the same results as shown in Table \ref{table:classification}.
Convolutional neural networks (CNN) experiments also used different number of layers and different number of filters and different filter sizes in each layer. The detailed structure of each layer in each CNN is described in Table \ref{table:models}. Similar values were found in terms of the accuracy and F1-score in all CNNs, together with small differences in the precision and recalls for both classes using CNNs with different structures as shown in Table \ref{table:models}. To train the FNN and CNN models, the binary cross entropy cost function and Adam optimizer were used. Because the initialization of the FNNs and CNNs is important, each model was trained three times with different random initialization for each training, i.e., training three models with the same structure but with different initializations.
\begin{table}[ht]
\centering
\caption{\small{Summary of the performance of the ML models for predicting normal middle ear condition and OME.}}
\label{table:classification}
\scalebox{0.8}
{
\begin{tabular}{|c|c|c|c|c|c|c|c|c|c|}
\hline
\multirow{2}{*}{Classifier} & \multirow{2}{*}{Design} &\multirow{2}{*}{AUC - ROC} &\multicolumn{2}{c|}{Precision} &\multicolumn{2}{c|}{Recall} &\multicolumn{2}{c|}{F1-score} &\multirow{2}{*}{Accuracy}   \\
                            &                         &                           &      Normal & OME             &   Normal & OME             &   Normal & OME               &                             \\
\hline
\multirow{3}{*}{KNN}        & 1                       &           0.74            &    0.83     & 0.64            &   0.75 & 0.73             &   0.79 & 0.68                 &      0.75                  \\
                            & 3                       &           0.75            &    0.81     & 0.68            &   0.81 & 0.69             &   0.81 & 0.68                 &      0.76                  \\
                            & 15                      &           0.78            &   0.81      & 0.80            &   0.91 & 0.65             &   0.86 & 0.72                 &      0.81                   \\
\hline
\multirow{4}{*}{SVM}        & Linear                  &           0.74            &   0.83      & 0.63            &   0.75 & 0.73             &   0.79  & 0.68                &   0.74                    \\
                            & Poly                    &           0.77            &   0.81      & 0.77            &   0.89 & 0.65             &   0.85  & 0.71                &   0.80                     \\
                            & RBF                     &           0.77            &  0.80       & 0.81            &   0.92 & 0.62             &   0.86  & 0.70                &   0.80                      \\
                            & Sigmoid                 &           0.74            &  0.79       & 0.72            &   0.86  & 0.62            &   0.82  & 0.66                &   0.77                       \\
\hline
\multirow{3}{*}{RF}         & 10                      &           0.77            &      0.81   & 0.75            &   0.87  & 0.66             &   0.84  & 0.70               &  0.79                       \\
                            & 100                     &           0.78            &      0.83   & 0.76            &   0.87  & 0.69             &   0.85  & 0.72               &  0.80                        \\
                            & 500                     &           0.78            &  0.83       & 0.76            &   0.87  & 0.69             &   0.85  & 0.72              &   0.80                      \\
\hline
\multirow{2}{*}{FNN}        & FNN1                    &           0.78            &   0.83      & 0.77            &   0.88  & 0.69             &   0.85  & 0.73              &  0.81                       \\
                            & FNN2                    &           0.79            &   0.83      & 0.78           &   0.89   & 0.69           &   0.86    & 0.73              &  0.81                       \\
\hline                            
\multirow{2}{*}{CNN}        & CNN1                    &           0.79            &   0.83     &  0.79          &   0.90 & 0.69             &   0.86     & 0.74              &  \textbf{0.82}              \\
                            & CNN2                    &            0.79           &   0.83     & 0.78           &   0.89 & 0.70             &   0.86     & 0.74            &  \textbf{0.82}                             \\         
\hline
\end{tabular}
}
\end{table}

\begin{table}[ht]
\centering
\caption{\small{The structures for the FNN and CNN models. The symbol “–”means the layer does not exist and Dense(1000) means fully connected layer with 1000 nodes. Conv2D (20, (21, 11)) means a 2D convolutional layer with 20 filters, where the size of each filter is 21 in the frequency direction and 11 in the pressure direction. MaxPooling2D (3,2), is a 2D max-polling operator with size 3 in the frequency direction and 2 in the pressure direction. Dropout (0.2) means 20\% dropout. 'relu' means rectified linear unit activation function.}}
\label{table:models}
\scalebox{0.8}
{
\begin{tabular}{|c|c|c||c|c|}
\hline
Layers / models                      &          FNN1                          &          FNN2                      &         CNN1                  &      CNN2   \\
\hline
\multirow{5}{*}{1}                   &  \multirow{2}{*}{Dense(1000)}          &  \multirow{2}{*}{Dense(1000)}      &    Conv2D(20, (21, 11))       &  Conv2D(20, (21, 11))   \\
                                     &                                        &                                    &    MaxPooling2D(3,2)          &  MaxPooling2D(3,2)   \\
                                     &\multirow{3}{*}{Activation(`relu')}     &\multirow{3}{*}{Activation(`relu')} &   BatchNormalization          &  BatchNormalization   \\
                                     &                                        &                                    &   Activation(`relu')          &  Activation(`relu')   \\
                                     &                                        &                                    &    Dropout(0.2)               &  Dropout(0.2)   \\
\hline
\multirow{4}{*}{2}                   & \multirow{2}{*}{Dense(100)}            & \multirow{2}{*}{Dense(500)}        & \multirow{4}{*}{Flatten}      &  Conv2D(40, (11, 7))   \\
                                     &                                        &                                    &                               &  BatchNormalization   \\
                                     &\multirow{2}{*}{Activation(`relu')}     &\multirow{2}{*}{Activation(`relu')} &                               &  Activation(`relu')   \\
                                     &                                        &                                    &                               &  Dropout(0.2)        \\
\hline
\multirow{4}{*}{3}                   & \multirow{2}{*}{Dense(1)}             & \multirow{2}{*}{Dense(100)}         &    Dense(100)                 &  Conv2D(60, (3, 3))   \\
                                     &                                       &                                     &    Activation(`relu')         &  BatchNormalization   \\
                                     &\multirow{2}{*}{Activation(`sigmoid')} &\multirow{2}{*}{Activation(`relu')}  &    Dropout(0.2)               &  Activation(`relu')   \\
                                     &                                       &                                     &                               &  Dropout(0.2)        \\
\hline
\multirow{2}{*}{4}                   &  \multirow{2}{*}{-}                   & Dense(1)                            &         Dense(1)              &  \multirow{2}{*}{Flatten}   \\
                                     &                                       & Activation(`sigmoid')               &   Activation(`sigmoid')       &                                \\
\hline                            
\multirow{3}{*}{5}                   &  \multirow{3}{*}{-}                   & \multirow{3}{*}{-}                  &   \multirow{3}{*}{-}          &  Dense(100)   \\
                                     &                                       &                                     &                               &  Activation(`relu')   \\
                                     &                                       &                                     &                               &  Dropout(0.25)   \\
\hline
\multirow{2}{*}{6}                   & \multirow{2}{*}{-}                    &  \multirow{2}{*}{-}                 &  \multirow{2}{*}{-}           &  Dense(1)   \\
                                     &                                       &                                     &                               &  Activation(`sigmoid')   \\                      
\hline
Number of parameters                 &  5,558,201                            &   6,008,701                         &   1,754,921                   & 5,338,621   \\
\hline
\end{tabular}
}
\end{table}

\subsection{Handling imbalanced datasets and the performance of the CNN models}
\begin{figure}[ht]
\centering
\includegraphics[width=0.8\linewidth, height=0.3\textheight]{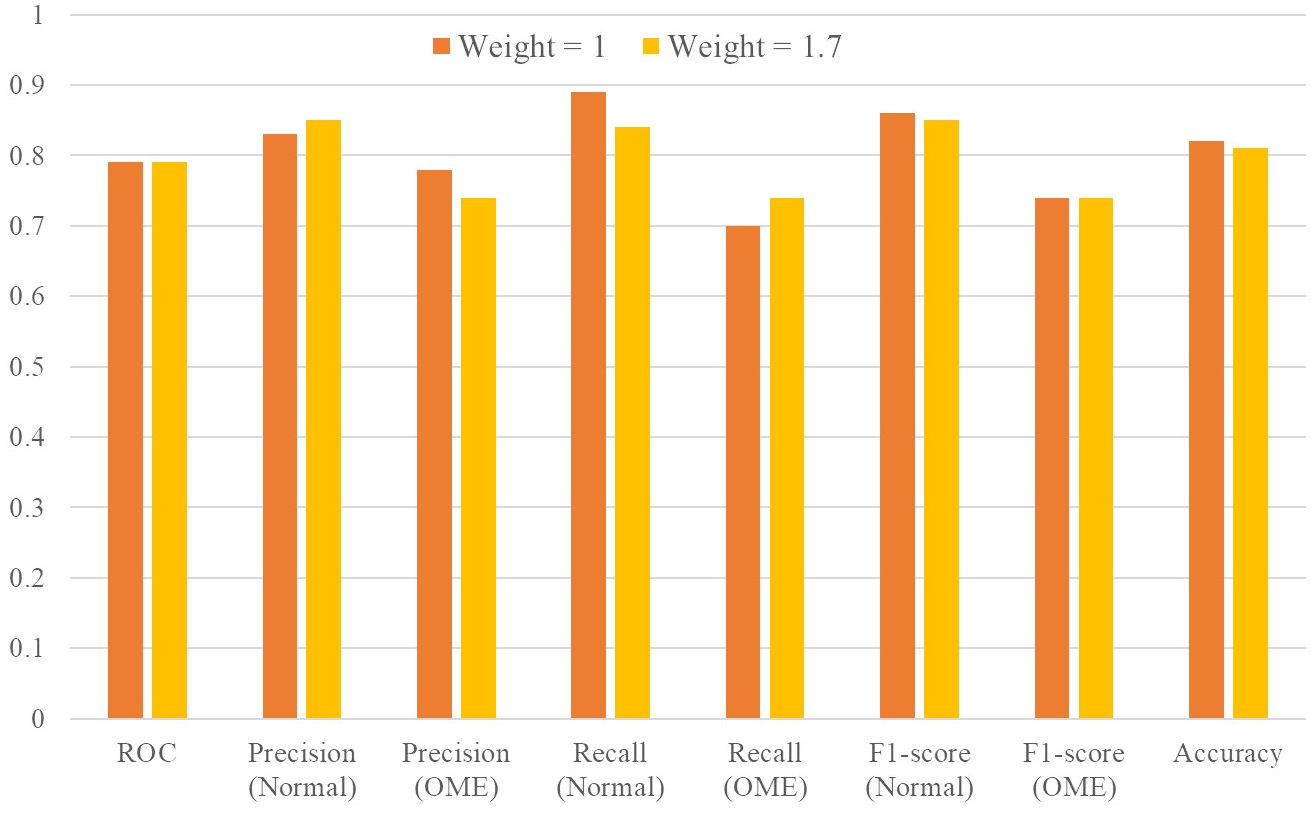}
\caption{\small{Performance of CNN2 with using different weights on the OME samples in the cost function}}
\label{fig3}
\end{figure}

As the dataset used in this study includes 423 samples from normal class and 249 samples from the OME class, i.e., the normal class samples are 1.7 times the OME samples, imbalance issue between the two classes may lead the classifiers to perform better with the majority class than the minority class \cite{Han:12:dmct,Sun:09:cidr}. This can be fixed using different methods, such as, resampling data space and cost-sensitive learning \cite{Sun:09:cidr}. In this study, a succinct approach based on cost-sensitive learning was used to penalize the errors arising from the misclassification of the minority class more than the error coming from the majority class during training the models. This was achieved by putting more weight on the errors from the misclassification of the minority class in the cost function for training the models \cite{Sun:09:cidr}. Consequently the error coming from the misclassification of the OME sample was weighted 1.7 times to the error coming from the misclassification of the samples from the normal middle ear class during training of the CNN models. The influence of the weighting rates was examined using the CNN2 model as shown in Table \ref{table:models}. Figure \ref{fig3} summarises the results obtained from using different weights to fix the imbalanced issue between normal and OME samples. The results showed an increase in the recall of the OME by 5\%, when we penalized the misclassification of the OME class by increasing the weight to 1.7. It implies that there is the possibility of improving predictive performance for OME cases by fixing the imbalance issue of the dataset.   
\subsection{Extracting the discriminative regions of the 2D frequency-pressure WAI images}
In this experiment, ML tools were used as data driven approaches to extract the discriminative regions from the WAI, which would provide useful guidance for clinicians in their diagnostic decisions. Two different data driven approaches were investigated:-
\begin{enumerate}
  \item   The first approach was based on extracting the key regions that derive the random forest (RF) classifier to categorize the WAI as normal or OME. In this approach, the RF classifier was tested as a feature selection approach to extract the key regions from the WAI;
  \item The second approach was to extract the regions in the 2D WAI images that indicate significant difference values between the classes using a  Wilcoxon rank sum test \cite{Hollander:15:nsm}.
\end{enumerate}
\subsubsection{Using random forest classifiers to extract the discriminative regions in the 2D WAI images}
The RF based feature selection technique extracts regions that mostly drive the classification decision by giving the important features that carry the most discriminative information more values than redundant features. Ten different RF with a different number of decision trees were tested, i.e., 10, 20, 30, 40, 50, 100, 200, 300, 400, and 500 trees. In each RF case, a set of coefficients were obtained representing the importance of the features in the 2D images. The coefficients from the ten RFs were then averaged to achieve a smoother estimate for the extracted regions. Figure \ref{fig4} shows the averaged coefficients from the ten RF cases. The highlighted regions with high values indicate discriminative regions for the different classes. There was an important discriminative region with high values around frequencies from 1000 Hz to 2670 Hz and pressures from -50 to +100 daPa that showed a significant difference between normal and OME data. 
\subsubsection{Using statistical significance tests to extract discriminative regions in the 2D WAI images}
\begin{figure}[ht]
  \centering
  \begin{minipage}[b]{0.49\textwidth}
    \includegraphics[width=\textwidth]{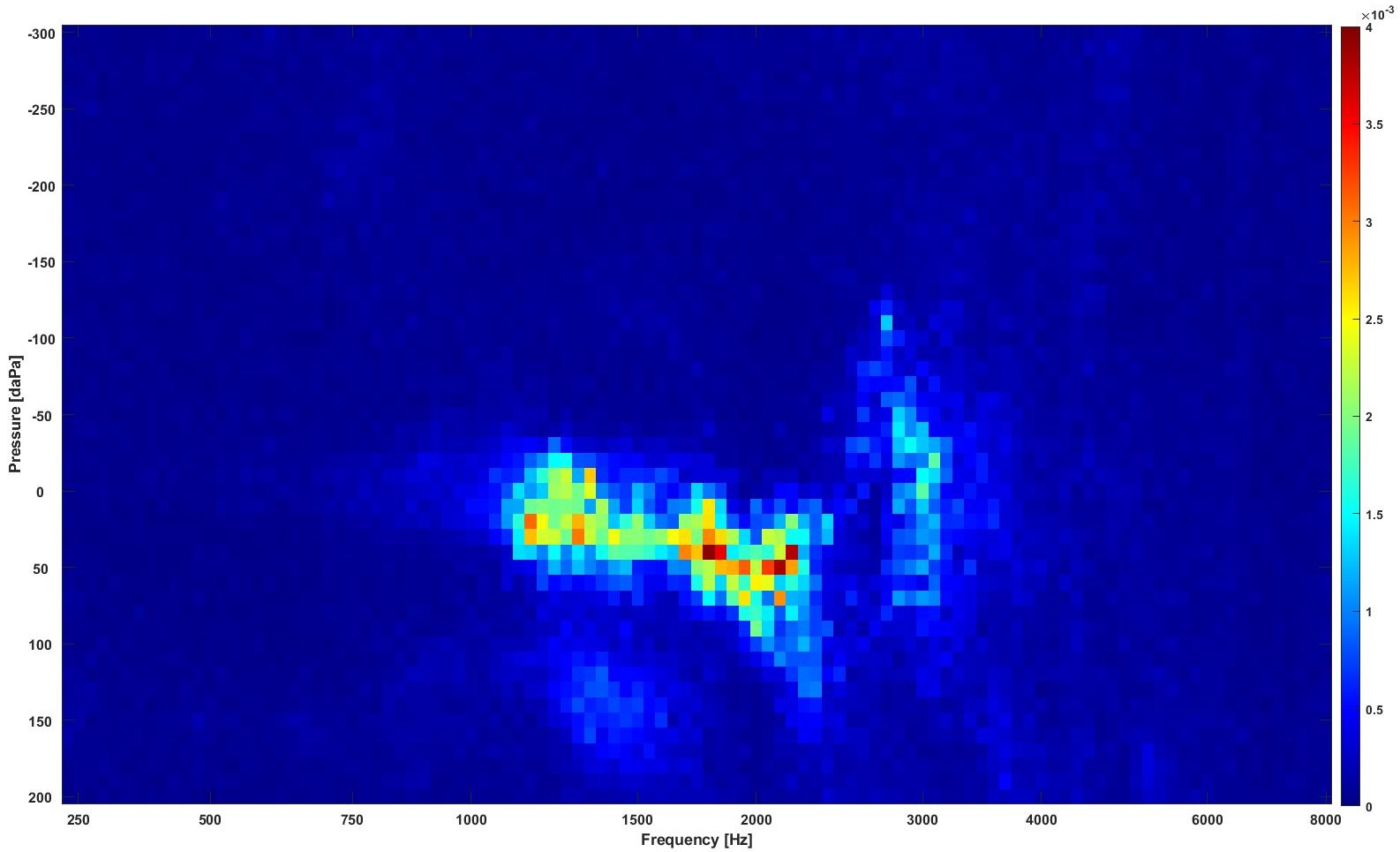}
    \caption{\small{The extracted discriminative regions in the WAI data using the RF classifiers. Regions with high values (bright regions) are the most discriminative regions.}}
    \label{fig4}
  \end{minipage}
  \hfill
  \begin{minipage}[b]{0.49\textwidth}
    \includegraphics[width=\textwidth]{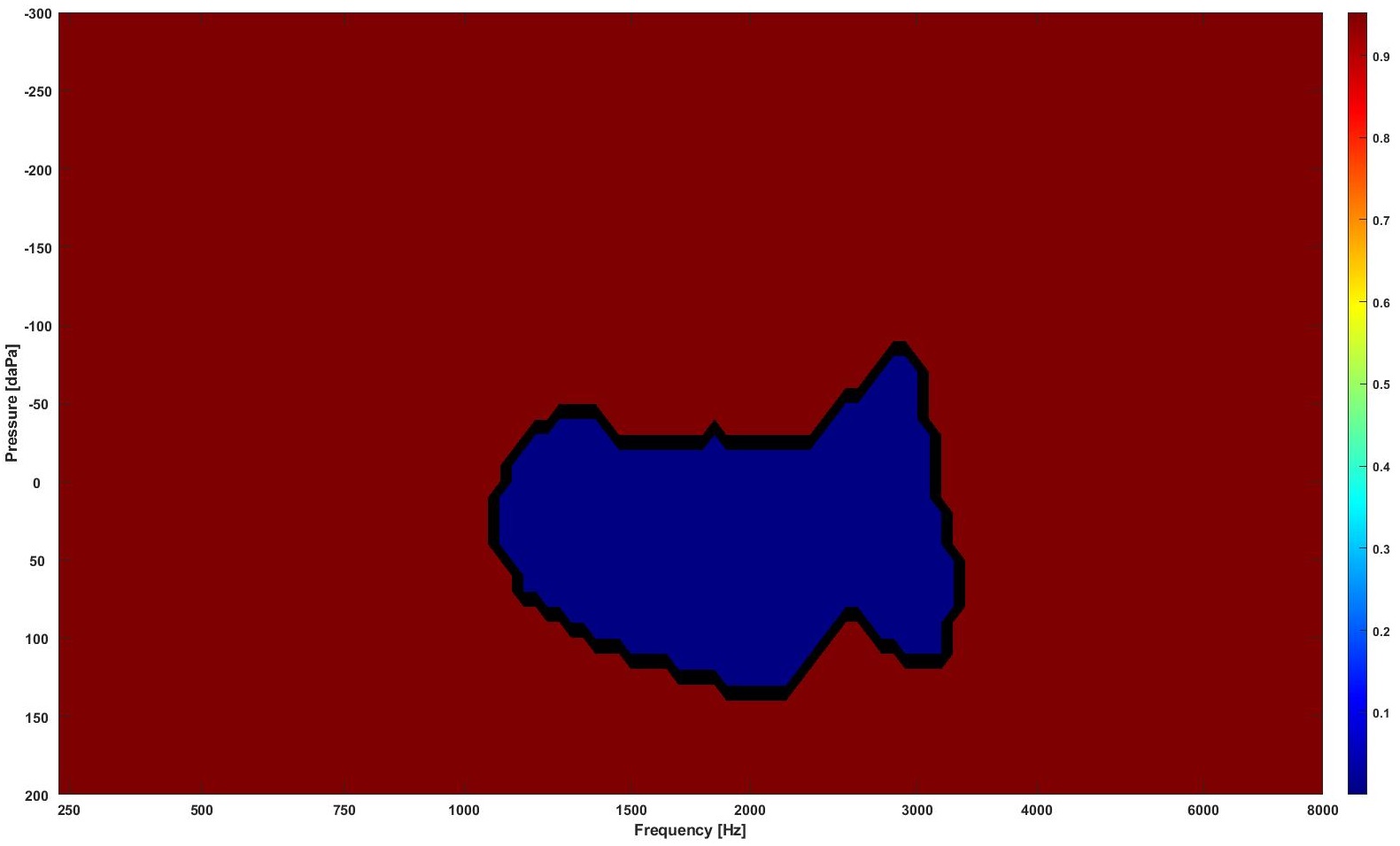}
    \caption{\small{The most significantly different region (blue) in the WAI data. The blue region is the region with the lowest $p$-values that contains 10\% of the data ($51*107*10\% = 546$ points).}}
    \label{fig5}
  \end{minipage}
\end{figure}

Section \ref{sec:waicsanmecceo} showed that 88\% of the total area in the 2D frequency-pressure images indicate significant difference in absorbance between normal and OME ears. In this experiment, the top 10\%, i.e., 5457 points 10\% = 546 points) of the most significantly different points, i.e., the lowest $p$-values between the two classes in the 2D frequency-pressure images were highlighted as shown in Figure \ref{fig5}. 

\begin{figure}[ht]
\centering
\includegraphics[width=0.80\linewidth, height=0.35\textheight]{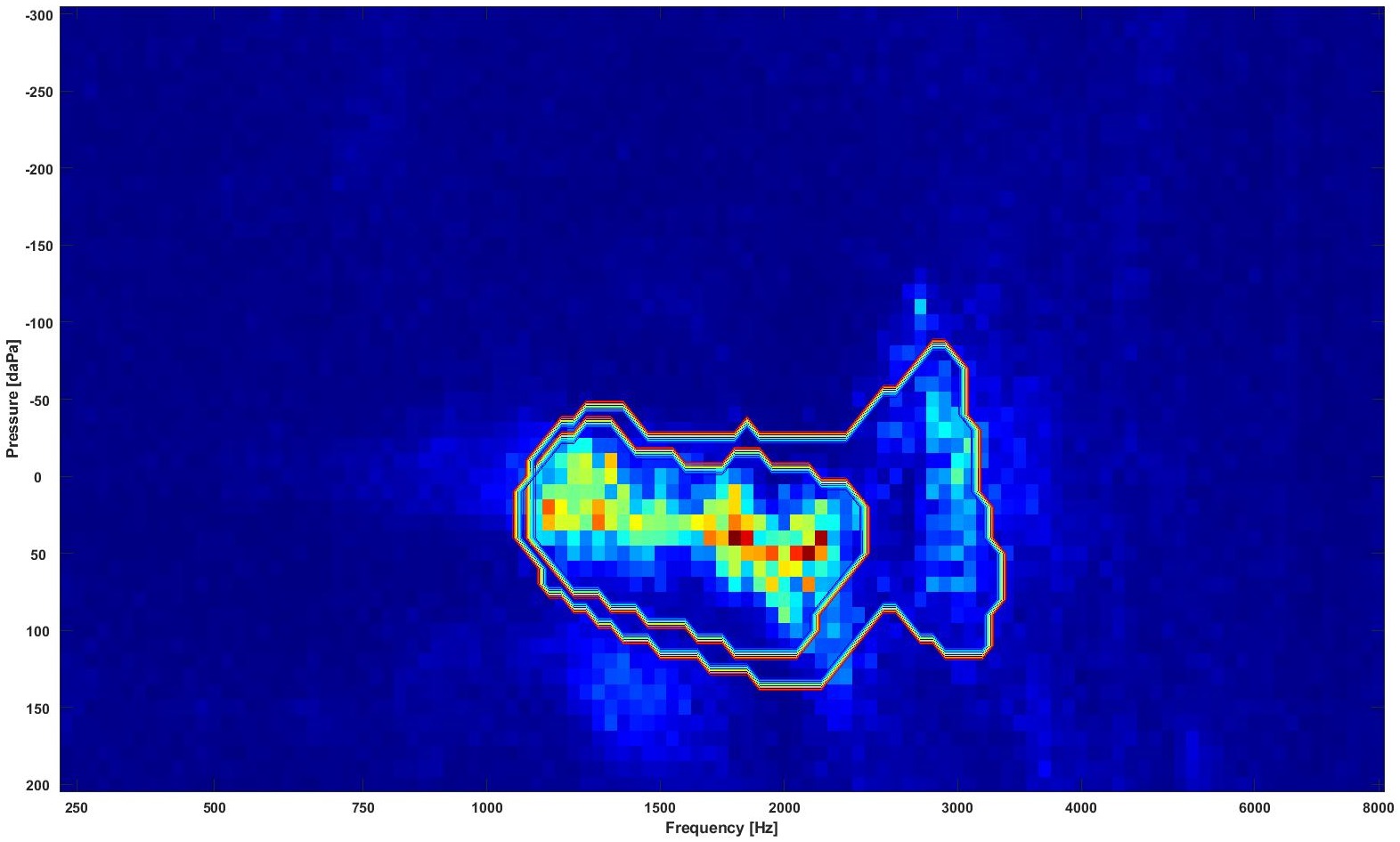}
\caption{\small{The combination of the regions extracted by RF and contours of extracted regions by the significance test, where the inner contour contains the 5\% most significant points (273 points) and the outer contour the top 10\% most significant points. Bright regions are the key regions extracted by the RF. The inner circle highlights the statistically significant area that contains 273 points with the lowest $p$-values (273/5457, 5.0\%). The outer circle features the statistically significant area that is comprised of points with the 546 lowest $p$-values (546/5457, 10.0\%).}}
\label{fig6}
\end{figure}

Figure \ref{fig6} shows regions extracted from RF with extracted region contours identified using the statistical significance test. The inner contour contains 5\% of the most significantly different points (273 points), the outer contour contains the top 10\% of the most significantly different points. Figure \ref{fig6} demonstrates that the RF and the significance test approaches picked almost the same regions. Further analysis showed the averaged absorbance in the 10\% extracted region was 0.59 and 0.33 for normal and OME ears respectively. For the 5\% extracted region mean absorbance was 0.53 for normal and 0.28 for OME ears.

%

\section{Discussion and Future Study}
According to the recent ENT Elective Care Handbook, the biggest challenge for ENT specialists appears to be ‘unnecessary’ referrals by GPs that could be managed effectively in primary care. An updated Clinical Practice Guideline on Otitis Media with Effusion (2016) \cite{Rosenfeld:16:cpgome} points out that a low percentage of clinicians follow clinical practice guidelines in the  use of pneumatic otoscopy for diagnosis due to a lack of experience in handling the technical difficulties of the device. Consequently efforts are still needed in primary care settings to teach and promote accurate OME diagnosis. 

In the present study, several ML classifiers have been examined using different parameters for the purposes of categorising middle ear function as normal or OME on the basis of WAI data. As shown in Table \ref{table:classification}, results from most of the tested classifiers are promising with the accuracy of most of the classifiers at around 80\%. Indeed the results from the ML classifiers in this study exceed diagnostic performance in identifying normal and OME ear conditions in the primary care settings by General Practitioners (GP) or other healthcare professionals using traditional middle ear diagnostic tools \cite{Wang:18:3diawtneome}. A study by Lee et al. \cite{Lee:04:cdaomecsmdt} investigated the accuracy of traditional diagnostic tools for OME, such as pneumatic otoscopy, otomicroscopy, and tympanometry. Their results showed low specificity in diagnosing childhood OME, although pneumatic otoscopy is recommended as the gold standard for OME diagnosis. In addition, there were high percentages of false positive and false negative cases when the results obtained from traditional tympanometry were examined. Future research will focus on improving the performance of the CNNs in terms of achieving more accurate and reliable classification results. Performance could be improved by using either large datasets or advanced deep learning techniques such as transfer learning, data augmentation, and few-shot learning \cite{Chen:19:clfsc} to train the CNNs more efficiently with small datasets. The work by Feyjie et al \cite{Feyjie:20:ssfslmis} demonstrates the efficiency of using few-shot learning in the task of skin lesion segmentation. A review on the state-of-the-art data augmentation methods applied in the context of segmenting brain tumours from MRI indicates that data augmentation has become a main part of almost all deep learning methods for segmenting brain lesions \cite{Nalepa:19:dabtsar}. Moreover, transfer learning has become a useful approach for analysing medical imaging using DL \cite{MORID2021104115}.  

This is the first study to use ML models to better understand and interpret the clinical meanings of characterized WAI regions that are closely associated with middle ear transfer function and further facilitate its clinical application. Two feature selection methods (i.e., random forest and statistical significant tests) were used in this study to extract the key regions from the WAI. The extracted regions could guide the clinicians to decide whether the case is normal or OME.  Lai et al. \cite{Lai:17:asfsmdatle} investigated three other feature selection methods to extract the dominant features to distinguish cases with temporal lobe epilepsy (TLE) from healthy cases. They are independent sample $t$-test filtering, the sparse-constrained dimensionality reduction model (SCDRM), and the support vector machine-recursive feature elimination (SVM-RFE). By using Support vector machine (SVM) to determine abnormal brain regions in TLE, their results indicated that the SVM-RFE achieved the best results, followed by the SCDRM and the $t$-test.
More advanced deep learning tools such as attention mechanisms could also be used to extract the key regions in our future studies \cite{ Zhou:16:ldfdl}. With attention mechanism, the neural network can weight features by level of importance to the classification task, and use this weighting to help achieve classification with better accuracy. Guan et al. \cite{Guan:18:dlragcnntdc} showed that the performance of automated classification of thorax disease on the basis of chest X-ray images using attention mechanisms was improved in terms of accuracy by cropping out the discriminative parts of the image and classifying both the global image as well as the cropped portion together. 
The size of the key region extracted and driving the classification decision is approximately 5\% of the whole WAI image (as shown in Figure \ref{fig4}), i.e., around frequencies from 1000Hz to 2670 Hz and pressure from -50 to +100 daPa, approximately 5\% of the whole WAI images. This result provides important guidance to ENT physicians, audiologists and other healthcare professionals in terms of WAI data interpretations and subsequent diagnostic process for identifying middle ear diseases in the clinical setting. The small size of the key regions suggests that dimensionality reduction techniques could be used before classification to decrease the size of the data, allowing efficient computing, simplifying the complexity of the problem and possible improvement of results \cite{Alvarez:19:amlfsidcnd}. The study by Zhao et al. \cite{Zhao:07:tumemocer} analysed the characteristics of 2D WAI plot configurations in ears with normal middle ear function. The results highlighted that the frequency region with high absorbance; 1.1 kHz (SD: 0.3 kHz) appeared related to  resonances in the middle ear system, where  sound energy coming into the external ear canal is transmitted most efficiently into the cochlea \cite{Keefe:93:ecirchia}. A previous study by Beers et al. (2010) \cite{Beers:10:wrnccsaccome} found that the area of the ROC curve was 0.9 at frequencies between 800 Hz and 5 kHz, with the best result at 1.25 kHz. 96\% sensitivity and 95\% specificity were achieved at the absorbance cut-off value of 71.7\% in diagnosing childhood OME with WAI. Their results also imply the importance of areas around the middle ear resonance frequency. 
Furthermore, Zhao et al. \cite{Zhao:07:tumemocer} found another region with high absorbance in the high frequency region (Mean: 3.4 kHz, SD: 1.5 kHz). They suggested that this region might be associated with the external ear canal resonance and middle ear structure. A recent study by Jungeun et al. \cite{Won:20:aemeewaioct} concluded that the otitis media group with high viscosity effusion had significantly less absorbance from 2.74 to 4.73 kHz in comparison to the otitis media group with low viscosity effusion. In addition, the amount of middle ear effusion affected the absorbance at the frequencies from 1.92 to 2.37 kHz. However, because of the complexity of 3D measurement results obtained from WAI, very few have explored the ability of WAI to differentiate between normal middle ears and OME, although a pilot study by Wang et al. \cite{Wang:18:3diawtneome} investigated the dynamic characteristics of the middle ear system using 3D image analysis in ears with normal middle ear function and in the OME condition. They reported that the areas in the frequency range between 1.0 and 8.0 kHz with normal middle ear pressure appeared important in terms of distinguishing normal from OME. Absorbance in the high frequency region under high positive pressure were significantly decreased in ears with OME. 
In the present study, the contour of averaged absorbance in the frequency-pressure  plot in normal ears is generally consistent with the findings of Hougaard et al. \cite{Hougaard:20:swtadcnha}. The averaged absorbance increases from 50\% at 1.0 kHz to an absorbance peak point around 75\% at 3.5 kHz under positive pressures between +50 and +150 daPa, followed by a sharp decrease at higher frequencies (Figure \ref{fig2}a). Averaged absorbance in ears with OME were significantly lower than those in normal ear conditions (Figure \ref{fig2}c). 
In comparison to the variance found in the normal ear conditions (Figure \ref{fig2}b), significantly higher variances were found in absorbance in ears with OME around the frequency from 4.0 kHz to 6.0 kHz in  the positive pressure region (Figure \ref{fig2}d). Jungeun et al. (2020) \cite{Won:20:aemeewaioct} also suggested a large variance in absorbance between 2.0 and 5.0 kHz in ears with OME of various type and amount of effusion. In a theoretical analysis using the Finite Element model of the middle ear, Koike and Wada \cite{Koike:02:mhmefem} suggested that positive pressure in the middle ear cavity had a greater impact on sound transmission than negative pressure at frequencies beyond 1.5 kHz. Therefore, the area with greater variance at the region of high frequency and positive middle ear pressure should be used as an indicator of severity in the OME condition.  

\section{Conclusion}
In this work, the accuracy and the area under the ROC curve obtained from the basic ML models were around 75\% and 80\%, respectively. The convolutional neural networks show slightly more promising results than the other models. The promising results from this study indicate that the ML approach is a useful tool to help the non-specialist healthcare practitioner in providing an effective and accurate method for the automated diagnosis of OME. A region around frequencies between 1090Hz to 2310 Hz and pressures from -40 to +90 daPa extracted from the WAI by the RF classifiers and the statistical significant tests indicates important areas to identify differences between normal and ears with OME. The significance results provide clear guidance to the practitioners to better understand and interpret the WAI data, and further facilitate its clinical applications. Future studies will focus on analyzing more WAI data on various middle ear disorders, e.g., otosclerosis, chronic otitis media and tympanic membrane perforation, using more robust deep learning tools.

\section*{Data sharing statements }
No additional data are available. However, the original data that support the findings derived from this study can be requested by emailing fzhao@cardiffmet.ac.uk.

\section*{Funding sources}
This work is supported by Sêr Cymru III Enhancing Competitiveness Infrastructure Award (MA/KW/5554/19), Great Britain Sasakawa Foundation (5826), Cardiff Metropolitan University Research Innovation Award and The Global Academies Research and Innovation Development Fund.

\section*{Author contributions}
Emad M Grais (EMG), Xiaoya Wang (ZYW), Jie Wang (JW) have contributed equally to the development of this project, but EMG was the main researcher in the analysis of results and writing of the manuscript. Fei Zhao (FZ) and Haidi Yang (HDY) conceptualized the project and contributed to the study design and writing of the original draft. FZ and HDY are co-corresponding authors. Wen Jiang (WJ), Yuexin Cai (YXC), Lifang Zhang (LFZ) and Wenqing Lin (WQL) contributed to data collection. All authors read and approved the final version of the manuscript. FZ and JW were involved in data review and verification. 

\section*{Acknowledgment}
We would like to acknowledge Dr. Christopher Wigham for the proofreading.

\section*{Declaration of Interests}
None declared.

\footnotesize
\bibliography{sample.bib}

\begin{thebibliography}{10}
\urlstyle{rm}
\expandafter\ifx\csname url\endcsname\relax
  \def\url#1{\texttt{#1}}\fi
\expandafter\ifx\csname urlprefix\endcsname\relax\def\urlprefix{URL }\fi
\expandafter\ifx\csname doiprefix\endcsname\relax\def\doiprefix{DOI: }\fi
\providecommand{\bibinfo}[2]{#2}
\providecommand{\eprint}[2][]{\url{#2}}

\bibitem{Zhao:09:feametfrp}
\bibinfo{author}{Zhao, F.}, \bibinfo{author}{Koike, T.}, \bibinfo{author}{Wang,
  J.}, \bibinfo{author}{Sienz, H.} \& \bibinfo{author}{Meredith, R.}
\newblock \bibinfo{journal}{\bibinfo{title}{{Finite} element analysis of the
  middle ear transfer functions and related pathologies}}.
\newblock {\emph{\JournalTitle{Medical Engineering and Physics}}}
  \textbf{\bibinfo{volume}{31}}, \bibinfo{pages}{907--916}
  (\bibinfo{year}{2009}).

\bibitem{Margolis:02:tbpca}
\bibinfo{author}{Margolis, R.} \& \bibinfo{author}{Hunter, L.}
\newblock \bibinfo{journal}{\bibinfo{title}{{Tympanometry - }basic principles
  and clinical applications}}.
\newblock {\emph{\JournalTitle{Contemporary perspectives in hearing assessment.
  (Boston: Allyn and Bacon)}}}  (\bibinfo{year}{2002}).

\bibitem{Shahnaz:97:smtnoe}
\bibinfo{author}{Shahnaz, N.} \& \bibinfo{author}{Polka, L.}
\newblock \bibinfo{journal}{\bibinfo{title}{{Standard} and multifrequency
  tympanometry in normal and otosclerotic ears}}.
\newblock {\emph{\JournalTitle{Ear and Hearing}}}
  \textbf{\bibinfo{volume}{18}}, \bibinfo{pages}{326--341}
  (\bibinfo{year}{1997}).

\bibitem{Zhao:02:medcpo}
\bibinfo{author}{Zhao, F.} \emph{et~al.}
\newblock \bibinfo{journal}{\bibinfo{title}{{Middle} ear dynamic
  characteristics in patients with otosclerosis}}.
\newblock {\emph{\JournalTitle{Ear and Hearing}}}
  \textbf{\bibinfo{volume}{23}}, \bibinfo{pages}{150--158}
  (\bibinfo{year}{2002}).

\bibitem{Margolis:99:wrtna}
\bibinfo{author}{Margolis, R.~H.}, \bibinfo{author}{Saly, G.~L.} \&
  \bibinfo{author}{Keefe, D.~H.}
\newblock \bibinfo{journal}{\bibinfo{title}{{Wideband} reflectance tympanometry
  in normal adults}}.
\newblock {\emph{\JournalTitle{Journal of the Acoustical Society of America}}}
  \textbf{\bibinfo{volume}{106}}, \bibinfo{pages}{265--280}
  (\bibinfo{year}{1999}).

\bibitem{Liu:08:watpssdranh}
\bibinfo{author}{Liu, Y.} \emph{et~al.}
\newblock \bibinfo{journal}{\bibinfo{title}{{Wideband} absorbance tympanometry
  using pressure sweeps: system development and results on adults with normal
  hearing}}.
\newblock {\emph{\JournalTitle{Journal of the Acoustical Society of America}}}
  \textbf{\bibinfo{volume}{124}}, \bibinfo{pages}{3708--3719}
  (\bibinfo{year}{2008}).

\bibitem{Zhao:07:tumemocer}
\bibinfo{author}{Zhao, F.}, \bibinfo{author}{Meredith, R.},
  \bibinfo{author}{Wotherspoon, N.} \& \bibinfo{author}{Rhodes, A.}
\newblock \bibinfo{title}{{Toward} an understanding of middle ear mechanics
  using otoreflectance: The characteristics of energy reflectances}.
\newblock In \emph{\bibinfo{booktitle}{4th Symposium on Middle Ear Mechanics
  and Otology}}, \bibinfo{pages}{59--68} (\bibinfo{year}{2007}).

\bibitem{Keefe:12:waaapchlc}
\bibinfo{author}{Keefe, D.~H.}, \bibinfo{author}{Sanford, C.~A.},
  \bibinfo{author}{Ellison, J.~C.}, \bibinfo{author}{Fitzpatrick, D.~F.} \&
  \bibinfo{author}{Gorga, M.~P.}
\newblock \bibinfo{journal}{\bibinfo{title}{{Wideband} aural acoustic
  absorbance predicts conductive hearing loss in children}}.
\newblock {\emph{\JournalTitle{International journal of audiology}}}
  \textbf{\bibinfo{volume}{51}}, \bibinfo{pages}{880—891}
  (\bibinfo{year}{2012}).

\bibitem{Keefe:03:etpchloca}
\bibinfo{author}{Keefe, D.~H.},  \& \bibinfo{author}{Simmons, J.~L.}
\newblock \bibinfo{journal}{\bibinfo{title}{{Energy} transmittance predicts
  conductive hearing loss in older children and adults}}.
\newblock {\emph{\JournalTitle{Journal of the Acoustical Society of America}}}
  \textbf{\bibinfo{volume}{114}}, \bibinfo{pages}{3217--3238}
  (\bibinfo{year}{2003}).

\bibitem{Wang:18:3diawtneome}
\bibinfo{author}{Wang, J.} \emph{et~al.}
\newblock \bibinfo{title}{{3D} image analysis of wideband tympanometry in
  normal ears of otitis media with effusion}.
\newblock In \emph{\bibinfo{booktitle}{8th Symposium on Middle Ear Mechanics
  and Otology}} (\bibinfo{year}{2018}).

\bibitem{Hougaard:20:swtadcnha}
\bibinfo{author}{Hougaard, D.~D.}, \bibinfo{author}{Lyhne, N.~M.},
  \bibinfo{author}{Skals, R.~K.} \& \bibinfo{author}{Kristensen, M.}
\newblock \bibinfo{journal}{\bibinfo{title}{{Study} on wideband tympanometry
  and absorbance within a danish cohort of normal hearing adults}}.
\newblock {\emph{\JournalTitle{European Archives of Oto-Rhino-Laryngology}}}
  \textbf{\bibinfo{volume}{277}}, \bibinfo{pages}{1899--1905}
  (\bibinfo{year}{2020}).

\bibitem{Niemczyk:19:wtamoe}
\bibinfo{author}{Niemczyk, E.}, \bibinfo{author}{Lachowska, M.},
  \bibinfo{author}{Tataj, E.}, \bibinfo{author}{Kurczak, K.} \&
  \bibinfo{author}{Niemczyk, K.}
\newblock \bibinfo{journal}{\bibinfo{title}{{Wideband} tympanometry and
  absorbance measurements in otosclerotic ears}}.
\newblock {\emph{\JournalTitle{Laryngoscope}}} \textbf{\bibinfo{volume}{129}},
  \bibinfo{pages}{365--376} (\bibinfo{year}{2019}).

\bibitem{Han:12:dmct}
\bibinfo{author}{Han, J.}, \bibinfo{author}{Kamber, M.} \&
  \bibinfo{author}{Pei, J.}
\newblock \emph{\bibinfo{title}{{Data} Mining: Concepts and Techniques}}
  (\bibinfo{publisher}{Morgan Kaufmann}, \bibinfo{year}{2012}).

\bibitem{Goodfellow-et-al-2016}
\bibinfo{author}{Goodfellow, I.}, \bibinfo{author}{Bengio, Y.} \&
  \bibinfo{author}{Courville, A.}
\newblock \emph{\bibinfo{title}{{Deep} Learning}} (\bibinfo{publisher}{MIT
  Press}, \bibinfo{year}{2016}).

\bibitem{Fritsch:80:mpci}
\bibinfo{author}{Fritsch, F.~N.} \& \bibinfo{author}{Carlson, R.~E.}
\newblock \bibinfo{journal}{\bibinfo{title}{Monotone piecewise cubic
  interpolation}}.
\newblock {\emph{\JournalTitle{SIAM Journal on Numerical Analysis}}}
  \textbf{\bibinfo{volume}{17}}, \bibinfo{pages}{238--246}
  (\bibinfo{year}{1980}).

\bibitem{chollet2015keras}
\bibinfo{author}{Chollet, F.} \emph{et~al.}
\newblock \bibinfo{title}{Keras}.
\newblock \bibinfo{howpublished}{\url{https://keras.io}}
  (\bibinfo{year}{2015}).

\bibitem{scikit-learn}
\bibinfo{author}{Pedregosa, F.} \emph{et~al.}
\newblock \bibinfo{journal}{\bibinfo{title}{Scikit-learn: Machine learning in
  {P}ython}}.
\newblock {\emph{\JournalTitle{Journal of Machine Learning Research}}}
  \textbf{\bibinfo{volume}{12}}, \bibinfo{pages}{2825--2830}
  (\bibinfo{year}{2011}).

\bibitem{Sun:09:cidr}
\bibinfo{author}{Sun, Y.}, \bibinfo{author}{Wong, A. K.~C.} \&
  \bibinfo{author}{Kamel, M.~S.}
\newblock \bibinfo{journal}{\bibinfo{title}{{Classification} of imbalanced
  data: {a Review}}}.
\newblock {\emph{\JournalTitle{International Journal of Pattern Recognition and
  Artificial Intelligence}}} \textbf{\bibinfo{volume}{23}},
  \bibinfo{pages}{687--719} (\bibinfo{year}{2009}).

\bibitem{Hollander:15:nsm}
\bibinfo{author}{Hollander, M.}, \bibinfo{author}{Wolfe, D.~A.} \&
  \bibinfo{author}{Chicken, E.}
\newblock \emph{\bibinfo{title}{{Nonparametric} Statistical Methods}}
  (\bibinfo{publisher}{Wiley}, \bibinfo{year}{2015}).

\bibitem{Rosenfeld:16:cpgome}
\bibinfo{author}{Rosenfeld, R.~M.} \emph{et~al.}
\newblock \bibinfo{journal}{\bibinfo{title}{{Clinical} practice guideline:
  Otitis media with effusion (update)}}.
\newblock {\emph{\JournalTitle{Otolaryngol Head Neck Surg}}}
  \textbf{\bibinfo{volume}{154}}, \bibinfo{pages}{1--41}
  (\bibinfo{year}{2016}).

\bibitem{Lee:04:cdaomecsmdt}
\bibinfo{author}{Lee, D.~H.} \& \bibinfo{author}{Yeo, S.~W.}
\newblock \bibinfo{journal}{\bibinfo{title}{{Clinical} diagnostic accuracy of
  otitis media with effusion in children, and significance of myringotomy:
  Diagnostic or therapeutic?}}
\newblock {\emph{\JournalTitle{Journal of the Korean Medical Science}}}
  \textbf{\bibinfo{volume}{19}}, \bibinfo{pages}{739--743}
  (\bibinfo{year}{2004}).

\bibitem{Chen:19:clfsc}
\bibinfo{author}{Chen, W.~Y.}, \bibinfo{author}{Liu, Y.~C.},
  \bibinfo{author}{Kira, Z.}, \bibinfo{author}{Wang, Y. C.~F.} \&
  \bibinfo{author}{Huang, J.~B.}
\newblock \bibinfo{title}{{A closer} look at few-shot classification}.
\newblock In \emph{\bibinfo{booktitle}{arXiv:1904.04232}}
  (\bibinfo{year}{2019}).

\bibitem{Feyjie:20:ssfslmis}
\bibinfo{author}{Feyjie, A.~R.} \emph{et~al.}
\newblock \bibinfo{title}{{Semi-supervised} few-shot learning for medical image
  segmentation}.
\newblock In \emph{\bibinfo{booktitle}{arXiv:2003.08462}}
  (\bibinfo{year}{2020}).

\bibitem{Nalepa:19:dabtsar}
\bibinfo{author}{Nalepa, J.}, \bibinfo{author}{Marcinkiewicz, M.} \&
  \bibinfo{author}{Kawulok, M.}
\newblock \bibinfo{journal}{\bibinfo{title}{{Data} augmentation for brain-tumor
  segmentation: {A Review}}}.
\newblock {\emph{\JournalTitle{Frontiers in Computational Neuroscience}}}
  \textbf{\bibinfo{volume}{13}} (\bibinfo{year}{2019}).

\bibitem{MORID2021104115}
\bibinfo{author}{Morid, M.~A.}, \bibinfo{author}{Borjali, A.} \&
  \bibinfo{author}{{Del Fiol}, G.}
\newblock \bibinfo{journal}{\bibinfo{title}{A scoping review of transfer
  learning research on medical image analysis using imagenet}}.
\newblock {\emph{\JournalTitle{Computers in Biology and Medicine}}}
  \textbf{\bibinfo{volume}{128}}, \bibinfo{pages}{104115},
  \doiprefix\url{https://doi.org/10.1016/j.compbiomed.2020.104115}
  (\bibinfo{year}{2021}).

\bibitem{Lai:17:asfsmdatle}
\bibinfo{author}{Lai, C.}, \bibinfo{author}{Guo, S.}, \bibinfo{author}{Cheng,
  L.} \& \bibinfo{author}{Wang, W.}
\newblock \bibinfo{journal}{\bibinfo{title}{A comparative study of feature
  selection methods for the discriminative analysis of temporal lobe
  epilepsy}}.
\newblock {\emph{\JournalTitle{Frontiers in Neurology}}}
  \textbf{\bibinfo{volume}{8}} (\bibinfo{year}{2017}).

\bibitem{Zhou:16:ldfdl}
\bibinfo{author}{Zhou, B.}, \bibinfo{author}{Khosla, A.},
  \bibinfo{author}{Lapedriza, A.}, \bibinfo{author}{Oliva, A.} \&
  \bibinfo{author}{Torralba, A.}
\newblock \bibinfo{title}{Learning deep features for discriminative
  localization}.
\newblock In \emph{\bibinfo{booktitle}{IEEE Conference on Computer Vision and
  Pattern Recognition (CVPR)}} (\bibinfo{year}{2016}).

\bibitem{Guan:18:dlragcnntdc}
\bibinfo{author}{Guan, Q.} \emph{et~al.}
\newblock \bibinfo{title}{{Diagnose} like a {Radiologist}: Attention guided
  convolutional neural network for thorax disease classification}.
\newblock In \emph{\bibinfo{booktitle}{arXiv:1801.09927}}
  (\bibinfo{year}{2018}).

\bibitem{Alvarez:19:amlfsidcnd}
\bibinfo{author}{Álvarez, J.~D.}, \bibinfo{author}{Matias-Guiu, J.~A.},
  \bibinfo{author}{Cabrera-Martín, M.~N.}, \bibinfo{author}{Risco-Martín,
  J.~L.} \& \bibinfo{author}{Ayala, J.~L.}
\newblock \bibinfo{journal}{\bibinfo{title}{{An} application of machine
  learning with feature selection to improve diagnosis and classification of
  neurodegenerative disorders}}.
\newblock {\emph{\JournalTitle{BMC Bioinformatics}}}
  \textbf{\bibinfo{volume}{20}} (\bibinfo{year}{2019}).

\bibitem{Keefe:93:ecirchia}
\bibinfo{author}{Keefe, D.~H.}, \bibinfo{author}{Bulen, J.~C.},
  \bibinfo{author}{Arehart, K.~H.} \& \bibinfo{author}{Burns, E.~M.}
\newblock \bibinfo{journal}{\bibinfo{title}{{Ear‐canal} impedance and
  reflection coefficient in human infants and adults}}.
\newblock {\emph{\JournalTitle{Journal of the Acoustical Society of America}}}
  \textbf{\bibinfo{volume}{94}}, \bibinfo{pages}{2617--2638}
  (\bibinfo{year}{1993}).

\bibitem{Beers:10:wrnccsaccome}
\bibinfo{author}{Beers, A.~N.}, \bibinfo{author}{Shahnaz, N.},
  \bibinfo{author}{Westerberg, B.~D.} \& \bibinfo{author}{Kozak, F.~K.}
\newblock \bibinfo{journal}{\bibinfo{title}{{Wideband} reflectance in normal
  caucasian and chinese school-aged children and in children with otitis media
  with effusion}}.
\newblock {\emph{\JournalTitle{Ear and Hearing}}}
  \textbf{\bibinfo{volume}{31}}, \bibinfo{pages}{221--233}
  (\bibinfo{year}{2010}).

\bibitem{Won:20:aemeewaioct}
\bibinfo{author}{Won, J.} \emph{et~al.}
\newblock \bibinfo{journal}{\bibinfo{title}{{Assessing} the effect of middle
  ear effusions on wideband acoustic immittance using optical coherence
  tomography}}.
\newblock {\emph{\JournalTitle{Ear and Hearing}}}
  \textbf{\bibinfo{volume}{41}}, \bibinfo{pages}{811--824}
  (\bibinfo{year}{2020}).

\bibitem{Koike:02:mhmefem}
\bibinfo{author}{Koike, T.}, \bibinfo{author}{Wada, H.} \&
  \bibinfo{author}{Kobayashi, T.}
\newblock \bibinfo{journal}{\bibinfo{title}{{Modeling} of the human middle ear
  using the finite-element method}}.
\newblock {\emph{\JournalTitle{Journal of the Acoustical Society of America}}}
  \textbf{\bibinfo{volume}{111}}, \bibinfo{pages}{1306--1317}
  (\bibinfo{year}{2002}).

\end{thebibliography}

\end{document}